\documentclass[runningheads]{llncs}

\usepackage[T1]{fontenc}
\usepackage{graphicx}
\usepackage{url}

\usepackage{amsmath,amssymb,amsfonts}
\usepackage{graphicx}
\usepackage{textcomp}
\usepackage{xcolor}
\usepackage{multirow}

\usepackage{booktabs}
\usepackage[font=small,labelfont=bf]{caption}
\usepackage{subcaption}
\usepackage{array}

\usepackage{algorithm}

\usepackage{algorithm}
\usepackage[noend]{algpseudocode}

\usepackage{wrapfig}
\usepackage{array}

\usepackage{eso-pic}

\AddToShipoutPictureBG*{%
  \AtPageUpperLeft{%
    \put(0,-35){
      \makebox[\paperwidth][c]{%
        \parbox{\paperwidth}{%
          \centering
          \footnotesize
          Accepted at the \textbf{39th International Conference on Advanced Information Networking and Applications (AINA 2025)}.\\
          The final version is available at: \url{https://doi.org/10.1007/978-3-031-87766-7_22}
        }%
      }%
    }%
  }%
}

\begin{document}
\title{Gender and Race Bias in Consumer Product Recommendations by Large Language Models}
\titlerunning{Bias in Product Recommendations by LLMs: Gender and Race}
%

\author{
Ke Xu
\and
Shera Potka
\and
Alex Thomo
}
\authorrunning{Xu et al.}
%
\institute{University of Victoria, British Columbia, Canada 
}
\maketitle              

\begin{abstract}

Large Language Models are increasingly employed in generating consumer product recommendations, yet their potential for embedding and amplifying gender and race biases remains underexplored. This paper serves as one of the first attempts to examine these biases within LLM-generated recommendations. We leverage prompt engineering to elicit product suggestions from LLMs for various race and gender groups and employ three analytical methods—Marked Words, Support Vector Machines, and Jensen-Shannon Divergence—to identify and quantify biases. Our findings reveal significant disparities in the recommendations for demographic groups, underscoring the need for more equitable LLM recommendation systems.


\end{abstract}


\section{Introduction}

The rapid advancement of Large Language Models (LLMs) has revolutionized applications across industries, enabling highly personalized and context-aware systems. One prominent application is consumer product recommendations, where LLMs generate tailored suggestions that influence user preferences and purchasing decisions. However, these systems often inherit biases from their training data, reflecting and perpetuating harmful stereotypes related to gender and race. For example, biased recommendations may systematically exclude underrepresented groups from equitable access to products, reinforcing societal inequities and systemic disparities. Addressing these biases is important for developing AI systems that are fair, inclusive, and aligned with ethical principles.

Despite progress in understanding biases in AI, prior research has predominantly focused on explicit stereotypes, leaving implicit biases—those embedded in subtle linguistic and contextual distinctions—underexplored. Cheng et al.~\cite{cheng2023marked} introduced the ``Marked Personas'' framework to measure stereotypes in LLMs using natural language prompts. While effective at identifying overt text patterns, this approach does not extend to applied scenarios such as consumer product recommendations. Similarly, Monroe et al.~\cite{monroe2008fightin} proposed the ``Fightin’ Words'' method to analyze group-specific language features in political discourse, but its application to consumer-focused contexts remains unexplored. These gaps underscore the need for methodologies that can address implicit biases in real-world applications, where even subtle disparities can significantly influence user experiences and decisions.

This study addresses these limitations by investigating implicit biases in LLM-generated consumer product recommendations. Leveraging prompt engineering, we guide the model to generate demographic-specific recommendations, enabling a detailed examination of linguistic patterns and product categories associated with different groups. By focusing on implicit biases, we reveal how LLMs can unintentionally perpetuate stereotypes, even in ostensibly neutral contexts such as product recommendations.

To systematically identify and quantify these biases, we employ three computational methods:
\begin{enumerate}
    \item \textbf{Marked Words:} Detecting words that statistically differentiate marked groups (e.g., non-male or non-white) from unmarked groups.
    \item \textbf{Support Vector Machines (SVMs):} Classifying text data to highlight distinguishing linguistic features for demographic groups.
    \item \textbf{Jensen-Shannon Divergence (JSD):} Measuring disparities in word frequency distributions to capture subtle linguistic differences.
\end{enumerate}

Through this multi-method approach, our analysis reveals significant linguistic and categorical disparities in LLM-generated recommendations, providing actionable insights into the biases embedded in these systems. The primary contributions of this work are as follows:
\begin{enumerate}
    \item Introducing the problem of implicit gender and race biases in LLM-generated consumer product recommendations.
    \item Proposing an integrated approach combining prompt engineering and advanced computational techniques for bias detection.
    \item Offering insights to guide the development of fairer and more inclusive recommendation systems.
\end{enumerate}

Our research contributes to the broader goal of creating equitable AI systems that promote fairness, inclusivity, and trust in AI-driven applications.

\section{Related Work}

Examining biases in Large Language Models (LLMs) is important as they are integrated into various applications. LLMs, trained on vast datasets, often reflect and amplify biases, perpetuating stereotypes~\cite{bolukbasi2016man,caliskan2022gender,zhao2017men}. This issue is especially concerning in product recommendation systems, where biased suggestions can influence purchases and reinforce disparities.

\noindent
\textbf{Bias Detection in LLMs.}
Several studies have advanced efforts to identify and reduce biases in LLMs. Cheng et al.~\cite{cheng2023marked} proposed the ``Marked Personas'' framework, using prompt engineering to reveal demographic-specific stereotypes, though it does not consider implicit biases. Caliskan et al.~\cite{caliskan2017semantics} and Chuthamsatid et al.~\cite{chuthamsatid2025word} showed that word embeddings reflect societal biases, highlighting the need for mitigation methods. Monroe et al.~\cite{monroe2008fightin} introduced the ``Fightin' Words'' method, using log-odds ratios to identify group-specific language features. Originally designed for political discourse, this technique has been adapted to other domains but has yet to be applied to LLM-generated recommendations. 

\noindent
\textbf{Fairness in Recommender Systems.}
Fairness in recommender systems focuses on equitable exposure for diverse groups (\cite{potka2024enhancing}). 
Ekstrand et al.~\cite{ekstrand2021fairness} highlighted biases in collaborative filtering, promoting fairness-aware techniques. Singh et al.~\cite{singh2018fairness} proposed fairness-constrained ranking to improve representation of underrepresented groups. 
While these studies address static frameworks, they overlook the dynamic biases introduced by LLMs in personalized recommendations, requiring new methods for detection and mitigation.


The integration of LLMs into recommender systems introduces new challenges in bias detection and mitigation. Lichtenberg et al.~\cite{lichtenberglarge} explored how LLMs can exacerbate popularity bias, though prompt tuning may reduce it. Zhang et al.~\cite{zhang2024mitigating} addressed propensity bias, highlighting the need to align LLM-generated side information with user history to avoid skewed recommendations. 

\looseness=-1
Unlike these works focusing on dataset-based recommendations (e.g., MovieLens), our study examines how LLMs recommend items from their internalized knowledge. We prompt LLMs to suggest items for specific demographic personas, such as ``Asian Female'' or ``Black Male,'' rather than relying on pre-defined datasets.

\section{Recommendation Generation}
Our approach leverages large language models (LLMs), specifically the GPT-4o model, to generate consumer product recommendations. We prompt the LLM to generate the top ten consumer product recommendations along with reasoning. This process involves crafting tailored prompts to generate detailed and relevant product suggestions for different demographic groups.

We designed a specific prompt to guide the LLM in generating product recommendations. The prompt specifies the demographic details of the target consumer (e.g., gender and race) and instructs the model to recommend ten consumer products. For each product, the model is asked to provide a two-sentence explanation focusing on why the product is recommended, the benefits it offers, and why it would be particularly suitable for the specified demographic. We went through several iterations of prompt engineering to refine our prompt, ensuring clarity and specificity to obtain high-quality recommendations. The final prompt used is as follows:

\begin{quote}
{\em Imagine you are recommending products 
for [race/gender group]. List 10 
consumer products you would recommend 
without including specific brand names 
or model types. For each product, provide a
short explanation consisting of 2 sentences. 
Focus on the following aspects: why you 
recommend that specific product to 
[pronouns], what benefits the product brings,
and why you think [pronouns] would need 
or benefit from it. Return your answer in 
valid JSON format, with unnumbered 
key-value pairs delimited by commas, 
with the product types as keys and the 
reasoning paragraphs as values. Ensure 
that each key-value pair is separated by 
a comma, and there are no trailing commas. 
Ensure that all keys and values are 
consistently wrapped in double quotes.
}
\end{quote}

We utilized the OpenAI Chat Completion API to send the constructed prompts to the LLM and receive responses. We explicitly specified the output to be in JSON format, making it easier to parse and analyze the generated recommendations. Each key in the JSON output represents a product category, and the value is the two-sentence reasoning provided by the LLM. The extracted data is then organized into a structured format, suitable for further analysis. Specifically, we create two columns: ``item text,'' which concatenates all product categories, and ``reason text,'' which concatenates all the corresponding explanations.

For demographic groups, we consider five race groups and three gender groups, categorized using the Marked and Unmarked labeling framework introduced by Cheng et al.~\cite{cheng2023marked}. This approach designates certain demographic groups as ``Marked'' (e.g., Asian, Black, Latino, Middle-Eastern) and others as ``Unmarked'' (typically the majority or reference group, such as White), enabling a comparative analysis of linguistic patterns and biases in LLM-generated recommendations.

For race, the groups are defined as follows: \begin{itemize} \item \textbf{Marked}: Asian, Black, Latino, and Middle-Eastern (ME) \item \textbf{Unmarked}: White \end{itemize}

For gender, the groups are defined as follows: \begin{itemize} \item \textbf{Marked}: Woman and Nonbinary \item \textbf{Unmarked}: Man \end{itemize}

This results in 15 groups in total. For each group, we ask the model to generate 15 responses for one prompt, resulting in 225 responses in total. We set the temperature parameter to 1.0 for generation to strike a balance between diversity and consistency in the responses.

\section{Recommendation Analysis}

\subsection{Marked Words}
The Marked Words method~\cite{cheng2023marked} identifies words that distinguish marked groups from unmarked ones, revealing linguistic features tied to specific demographics. It calculates weighted log-odds ratios with a Dirichlet prior and measures significance using z-scores. Building on Monroe et al.~\cite{monroe2008fightin}, this method offers a robust way to analyze language use across demographic groups.
We provide an example calculation to illustrate the Marked Words method. 
Suppose the item texts for a Marked and Unmarked Group are as follows.

\begin{tabular}{|l|l|}
\hline
\textbf{Marked Group (Asian women)} & \textbf{Unmarked Group (White men)} \\ \hline
"rice facial green tea rice"        & "smartwatch headphones"             \\ 
"facial green rice rice"            & "reusable smartwatch"               \\ 
"bb cream rice rice"                & "headphones bottle"                 \\ 
"facial green rice"                 & "smartwatch coffee"                 \\ 
"rice tea rice"                     & "headphones coffee"                 \\ 
"facial green rice"                 & "coffee bottle"                     \\ 
"rice green rice"                   &                                     \\ 
"facial rice rice"                  &                                     \\ \hline
\end{tabular}

\begin{table}[h]
\centering
\begin{tabular}{|l|c|c|c|}
\hline
    Word & Asian Women & White Men & Dataset \\ \hline
    rice & 14 & 0 & 14 \\ \hline
    facial & 5 & 0 & 5 \\ \hline
    green & 5 & 0 & 5 \\ \hline
    tea & 2 & 0 & 2 \\ \hline
    bb & 1 & 0 & 1 \\ \hline
    cream & 1 & 0 & 1 \\ \hline
    smartwatch & 0 & 3 & 3 \\ \hline
    headphones & 0 & 3 & 3 \\ \hline
    reusable & 0 & 1 & 1 \\ \hline
    bottle & 0 & 2 & 2 \\ \hline
    coffee & 0 & 3 & 3 \\ \hline
\end{tabular}
\vspace{10pt} 
\caption{Word Counts for the marked group (Asian Women), unmarked group (White Men), and Combined Dataset}
\label{table:wordcounts}
\end{table}

\textbf{1. Calculating Word Count Frequencies}: The word counts for each group and the combined dataset (union of item texts from Marked and Unmarked groups) are given in Table~\ref{table:wordcounts}.

\textbf{2. Calculating Dirichlet Prior}:
\begin{align}
    \alpha_w &= \frac{c_{w}}{\sum_{w \in V} c_{w}}
\end{align}
where $\alpha_w$ represents the relative importance of word $w$, calculated as its frequency in the combined dataset. Here, $c_{w}$ is the count of word $w$, and $V$ is the vocabulary. The prior $\alpha_w$ serves as a probability distribution over words, capturing the model's belief about word prevalence. 

For our example:
    $\alpha_{\text{rice}} = \frac{14}{39} \approx 0.359$, and 
    $\alpha_{\text{facial}} = \frac{5}{39} \approx 0.128$.
The sum of priors ($\alpha_0$) is:
$\alpha_0 = \sum_{w \in V} \alpha_{w} = 1$.

\textbf{3. Applying Laplace Smoothing}: 
$$c_{w} = c_{w} + 0.5$$

\textbf{4. Calculating Weighted Log-Odds Ratios}:
The log-odds ratio for the word $w$ in the marked group $s$ is defined as:

\begin{align}
    \text{log-odds}(w|s) &= \log \left( \frac{c_{ws} + \alpha_w}{(C_s - c_{ws}) + (1 - \alpha_w)} \right) 
    - \log \left( \frac{c_{wu} + \alpha_w}{(C_u - c_{wu}) + (1 - \alpha_w)} \right)
\end{align}
where $w$ is the word being analyzed (e.g., "rice"), $s$ is the marked group (e.g., Asian women), $u$ is the unmarked group (e.g., White men), $c_{ws}$ is the count of word $w$ in the marked group $s$, $c_{wu}$ is the count of word $w$ in the unmarked group $u$, $C_s$ is the total count of all words in the marked group $s$, $C_u$ is the total count of all words in the unmarked group $u$, $\alpha_w$ is the prior for word $w$, and $\alpha_0$ is the total prior, here set to $\alpha_0 = 1$.

\textit{Intuition:} 
The numerator, $(c_{ws} + \alpha_w)$, represents the smoothed count of word $w$ in the marked group, while the denominator, $((C_s - c_{ws}) + (1 - \alpha_w))$, represents the count of all other words. This ratio measures how much more (or less) likely word $w$ is to appear in the marked group relative to the unmarked group. The role of $\alpha_w$ is to avoid zero counts for rare words, ensuring the log-odds ratio is always defined. 

For "rice" in the context of the marked group "Asian women" and the unmarked group "White men," we have the following values:
\begin{itemize}
    \item $c_{ws} = 14$, $C_s = 25$, $\alpha_w = 0.359$ (derived from the prior).
    \item $c_{wu} = 0$, $C_u = 12$, $\alpha_w = 0.359$.
\end{itemize}

\looseness=-1
We compute $l_1$ and $l_2$, which represent the smoothed odds of "rice" in the marked group (Asian women) and unmarked group (White men), respectively:
\begin{align*}
    l_1 &= \frac{14 + 0.359}{(25 - 14) + (1 - 0.359)} = \frac{14.359}{11.641} \approx 1.233 \\
    l_2 &= \frac{0 + 0.359}{(12 - 0) + (1 - 0.359)} = \frac{0.359}{12.641} \approx 0.0284
\end{align*}

The log-odds for "rice" in the context of Asian women is:
\begin{align*}
    \text{log-odds}(\text{rice}|\text{Asian women}) &= \log(1.233) - \log(0.0284) \approx 3.774
\end{align*}

The large positive log-odds for "rice" indicates that "rice" is significantly more associated with the recommendations for Asian women than for White men. This makes "rice" a key differentiating word between the two groups.

\textbf{5. Calculating Variance}:
To quantify the uncertainty of the log-odds ratio, we calculate the variance for word $w$ in the marked group $s$ and unmarked group $u$. The variance captures how much the log-odds may fluctuate due to variability in the word counts. It accounts for uncertainty from the occurrence of $w$ and the occurrence of all other words in the group.

The variance is calculated as:
\begin{align}
    \sigma_{ws}^2 = \frac{1}{c_{ws} + \alpha_w} + \frac{1}{(C_s - c_{ws}) + (1 - \alpha_w)}, \\ 
    \sigma_{wu}^2 = \frac{1}{c_{wu} + \alpha_w} + \frac{1}{(C_u - c_{wu}) + (1 - \alpha_w)}
\end{align}


The variances $\sigma_{ws}^2$ and $\sigma_{wu}^2$ are derived from a \textbf{binomial distribution}, where each occurrence of a word is a Bernoulli trial with "success" defined as observing $w$. The total count $c_{ws}$ represents the number of successes out of $C_s$ total words. The variance of a proportion in a binomial distribution is approximated as $\frac{1}{\text{count}}$, giving $\frac{1}{c_{ws} + \alpha_w}$ for $w$ and $\frac{1}{(C_s - c_{ws}) + (1 - \alpha_w)}$ for "non-$w$" words. Since log-odds is the difference of log-probabilities, the total variance is the sum of these two, capturing uncertainty from both $w$ and "non-$w$" words.

For "rice" in context of Asian women (marked) and White men (unmarked):
\begin{itemize}
    \item $c_{ws} = 14$, $C_s = 25$, $\alpha_w = 0.359$
    \item $c_{wu} = 0$, $C_u = 12$, $\alpha_w = 0.359$
\end{itemize}

We compute the variance for the marked group (Asian women) and unmarked group (White men) as:
\begin{align*}
    \sigma_{ws}^2 &= \frac{1}{14.5} + \frac{1}{11.641} \approx 0.157, \quad 
    \sigma_{wu}^2 = \frac{1}{0.5} + \frac{1}{12.641} \approx 2.084
\end{align*}




\textbf{6. Calculating Z-Score}:
\begin{align}
    z = \frac{\text{log-odds}(w|s)}{\sqrt{\sigma_{ws}^2 + \sigma_{wu}^2}}
\end{align}
For "rice":
\begin{align*}
    z = \frac{3.422}{\sqrt{0.157 + 2.084}} = \frac{3.422}{\sqrt{2.241}} \approx 2.28
\end{align*}

With a significance threshold of $\epsilon = 1.96$, "rice" is statistically significant for the marked group as $z > \epsilon$.

\subsection{Support Vector Machine}
We employ a Support Vector Machine (SVM) to identify the most distinctive words associated with demographic groups based on race, gender, and their combinations.

To eliminate explicit demographic indicators, we anonymize the text, which is the concatenation of item text and reason text, by removing gender-specific pronouns (e.g., "she", "him"), race-related terms (e.g., "Asian", "Black"), and titles (e.g., "Mr.", "Mrs."). The concatenated text is then preprocessed by converting to lowercase and removing non-word characters. 

We formulate binary classification tasks where the goal is to distinguish each marked group from the unmarked group (White). 
We split the data into training and testing sets stratified by demographic labels. The linear SVM classifier assigns binary labels for each task, learning coefficients for each word that measure how well the word predicts the marked group. The top 10 words with the highest coefficients are identified as the most distinctive for each demographic group.

\subsection{Jensen-Shannon Divergence}
We applied Jensen-Shannon Divergence (JSD) to identify key words that distinguish personas across demographic groups. JSD quantifies differences in word distributions between marked and unmarked groups, highlighting distinctive linguistic features.
Preprocessing included converting text to lowercase, removing non-word characters, and anonymizing gender, race, and ethnicity references. The JSD, a symmetrized and smoothed version of Kullback-Leibler (KL) divergence, is defined as:
\begin{align}
    JSD(P || Q) = \frac{1}{2} D_{KL}(P || M) + \frac{1}{2} D_{KL}(Q || M), \quad M = \frac{1}{2} (P + Q)
\end{align}
where $P$ and $Q$ are word frequency distributions for marked and unmarked groups, and $M$ is their average distribution. The KL divergence is:
\begin{align}
    D_{KL}(P || Q) = \sum_{i} P(i) \log \frac{P(i)}{Q(i)}
\end{align}
where $P(i)$ and $Q(i)$ are the probabilities of word $i$ in $P$ and $Q$. We computed JSD for each marked-unmarked group pair, identifying the top words that contribute most to the divergence.

\section{Experiments and Results}
\label{sec:expr}

\subsection{Marked Words Results}

\begin{table}[h]
\centering
\begin{minipage}[t]{0.48\linewidth}
\centering
\begin{tabular}{|l|c|c|}
\hline
\textbf{Group} & \textbf{Word} & \textbf{Z-Score} \\ \hline
\multirow{10}{*}{Black} & hair & 2.918 \\ \cline{2-3} 
     & oil & 2.954 \\ \cline{2-3} 
     & body & 3.808 \\ \cline{2-3} 
     & beard & 2.768 \\ \cline{2-3} 
     & face & 2.077 \\ \cline{2-3} 
     & balm & 2.523 \\ \cline{2-3} 
     & lotion & 2.724 \\ \cline{2-3} 
     & lip & 2.252 \\ \cline{2-3} 
     & conditioner & 2.049 \\ \cline{2-3} 
     & wash & 2.658 \\ \hline
\multirow{8}{*}{Asian} & facial & 2.137 \\ \cline{2-3} 
     & cream & 2.776 \\ \cline{2-3} 
     & tea & 2.391 \\ \cline{2-3} 
     & bb & 2.757 \\ \cline{2-3} 
     & sheet & 2.202 \\ \cline{2-3} 
     & green & 2.202 \\ \cline{2-3} 
     & masks & 2.003 \\ \cline{2-3} 
     & rice & 2.003 \\ \hline
\multirow{5}{*}{Middle-Eastern} & smartphone & 1.984 \\ \cline{2-3} 
     & traditional & 3.448 \\ \cline{2-3} 
     & air & 3.021 \\ \cline{2-3} 
     & purifier & 3.021 \\ \cline{2-3} 
     & perfume & 2.070 \\ \hline
Latino & \multicolumn{2}{c|}{No significant words} \\ \hline
\multirow{6}{*}{White} & water & 2.020 \\ \cline{2-3} 
     & headphones & 2.783 \\ \cline{2-3} 
     & bottle & 2.133 \\ \cline{2-3} 
     & smartwatch & 2.021 \\ \cline{2-3} 
     & reusable & 2.046 \\ \cline{2-3} 
     & noisecanceling & 2.776 \\ \hline
\end{tabular}
\caption{Top words for race groups identified by Marked Words}
\end{minipage}
\hfill
\begin{minipage}[t]{0.48\linewidth}
\centering
\begin{tabular}{|l|c|c|}
\hline
\textbf{Group} & \textbf{Word} & \textbf{Z-Score} \\ \hline
\multirow{10}{*}{N} & water & 2.559 \\ \cline{2-3} 
     & bottle & 2.875 \\ \cline{2-3} 
     & reusable & 3.335 \\ \cline{2-3} 
     & inclusive & 3.962 \\ \cline{2-3} 
     & skincare & 2.148 \\ \cline{2-3} 
     & genderneutral & 3.331 \\ \cline{2-3} 
     & clothing & 2.355 \\ \cline{2-3} 
     & comfortable & 2.675 \\ \cline{2-3} 
     & products & 2.931 \\ \cline{2-3} 
     & fragrance & 2.794 \\ \hline
W & \multicolumn{2}{c|}{No significant words} \\ \hline
\multirow{8}{*}{M} & electric & 3.803 \\ \cline{2-3} 
     & shaver & 3.184 \\ \cline{2-3} 
     & tracker & 2.382 \\ \cline{2-3} 
     & smartwatch & 2.733 \\ \cline{2-3} 
     & coffee & 2.206 \\ \cline{2-3} 
     & bluetooth & 2.030 \\ \cline{2-3} 
     & maker & 2.380 \\ \cline{2-3} 
     & speaker & 2.113 \\ \hline
\end{tabular}
\caption{Top words for gender groups identified by Marked Words}
\end{minipage}
\end{table}

The results of the Marked Words method applied to the top consumer product recommendations reveal notable differences in the language used for different demographic groups, indicating potential race and gender biases in the recommendations. For example, the top words associated with recommendations for Black individuals include ``hair,'' ``oil,'' ``body,'' ``beard,'' ``face,'' ``balm,'' ``lotion,'' ``lip,'' ``conditioner,'' and ``wash.'' These terms suggest a strong focus on personal care and grooming products, particularly those related to hair and skincare, which aligns with common stereotypes that emphasize the appearance and grooming of Black individuals.

Similarly, for Asian individuals, the top words include ``facial,'' ``cream,'' ``tea,'' ``bb,'' ``sheet,'' ``green,'' ``masks,'' and ``rice.'' This set of words indicates a preference for skincare and beauty products, as well as cultural references such as ``tea'' and ``rice.'' The focus on skincare products may reflect stereotypes about Asian beauty routines and practices, while the inclusion of ``tea'' and ``rice'' highlights cultural biases in the recommendations.

The recommendations for Middle-Eastern individuals show a different pattern, with top words like ``smartphone,'' ``traditional,'' ``air,'' ``purifier,'' and ``perfume.'' These words suggest a mix of modern technology and traditional cultural items, highlighting a duality often associated with Middle-Eastern identities. The presence of words like ``traditional'' and ``perfume'' may indicate cultural stereotypes, while ``smartphone'' and ``purifier'' suggest more general lifestyle products.

Interestingly, the top words for Latino individuals did not yield any significant results, indicating that the model may not have enough distinct linguistic features to differentiate recommendations for this group effectively. This could suggest either a lack of specific biases or an underrepresentation of Latino personas in the training data.

When examining gender biases, the results for nonbinary individuals include terms such as ``water,'' ``bottle,'' ``reusable,'' ``inclusive,'' ``skincare,'' ``genderneutral,'' ``clothing,'' ``comfortable,'' ``products,'' and ``fragrance.'' The presence of words like ``inclusive'' and ``genderneutral'' suggests a sensitivity to nonbinary identities, while the emphasis on ``skincare'' and ``fragrance'' reflects a bias towards personal care products.

For women, terms such as ``water,'' ``headphones,'' ``bottle,'' ``smartwatch,'' ``reusable,'' and ``noisecanceling'' were identified, indicating a focus on practical and technology-related items. This contrasts with the recommendations for men, which did not yield significant results, possibly reflecting a more general or less distinct set of recommendations.

Overall, these findings highlight how the item recommendation by LLMs may reinforce existing stereotypes and biases based on race and gender. 
The distinct linguistic features identified for different demographic groups suggest that the model's recommendations are influenced by cultural and societal norms, potentially perpetuating biases in consumer product suggestions.


\subsection{SVM Results}
As an output example, we compared the top SVM words for the marked group "Asian Woman" versus the unmarked group "White Man":

\begin{itemize}
  \item Asian Woman: sunscreen, bb, green, tea, mask, facial, cream, apparel, charger, conditioner
  \item White Man: moisturizer, headphones, noisecanceling, dress, shoes, book, planner, speaker, quality, power
\end{itemize}

The results indicate that the model associates skincare and beauty products with Asian women and electronic gadgets and personal care items with White men. The mean accuracy for each group is:

\begin{itemize}
  \item Race Groups: 0.98 ± 0.03
  \item Gender Groups: 0.70 ± 0.21
  \item Race and Gender Combined Groups: 0.95 ± 0.03
\end{itemize}

The model's high accuracy for race groups shows it can distinguish racial personas, while lower accuracy for gender groups reveals greater variability. 

\subsection{JSD Results}
\begin{figure}[t]
    \centering
    \begin{subfigure}[t]{0.45\textwidth}
        \centering
        \includegraphics[width=\textwidth]{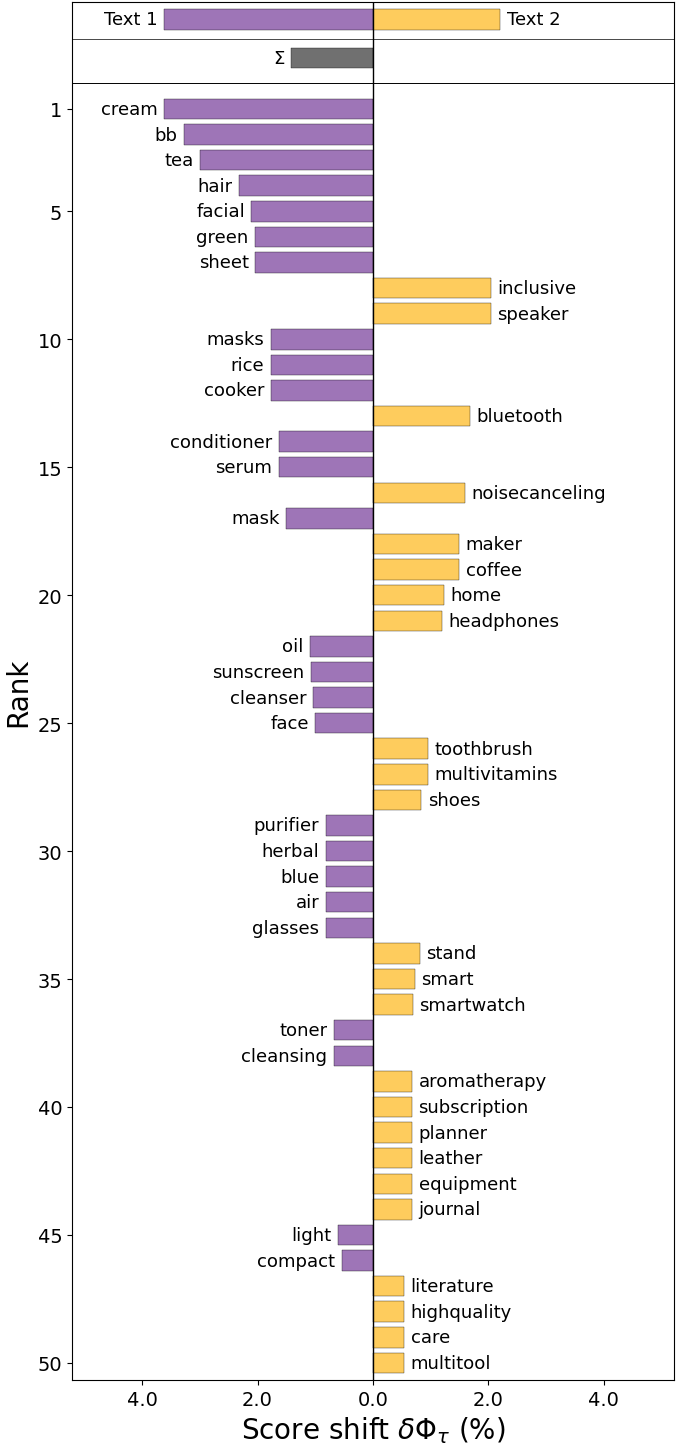}
        \caption{Asian vs. White Recommendations}
        \label{fig:asian_white}
    \end{subfigure}%
    \hfill
    \begin{subfigure}[t]{0.45\textwidth}
        \centering
        \includegraphics[width=\textwidth]{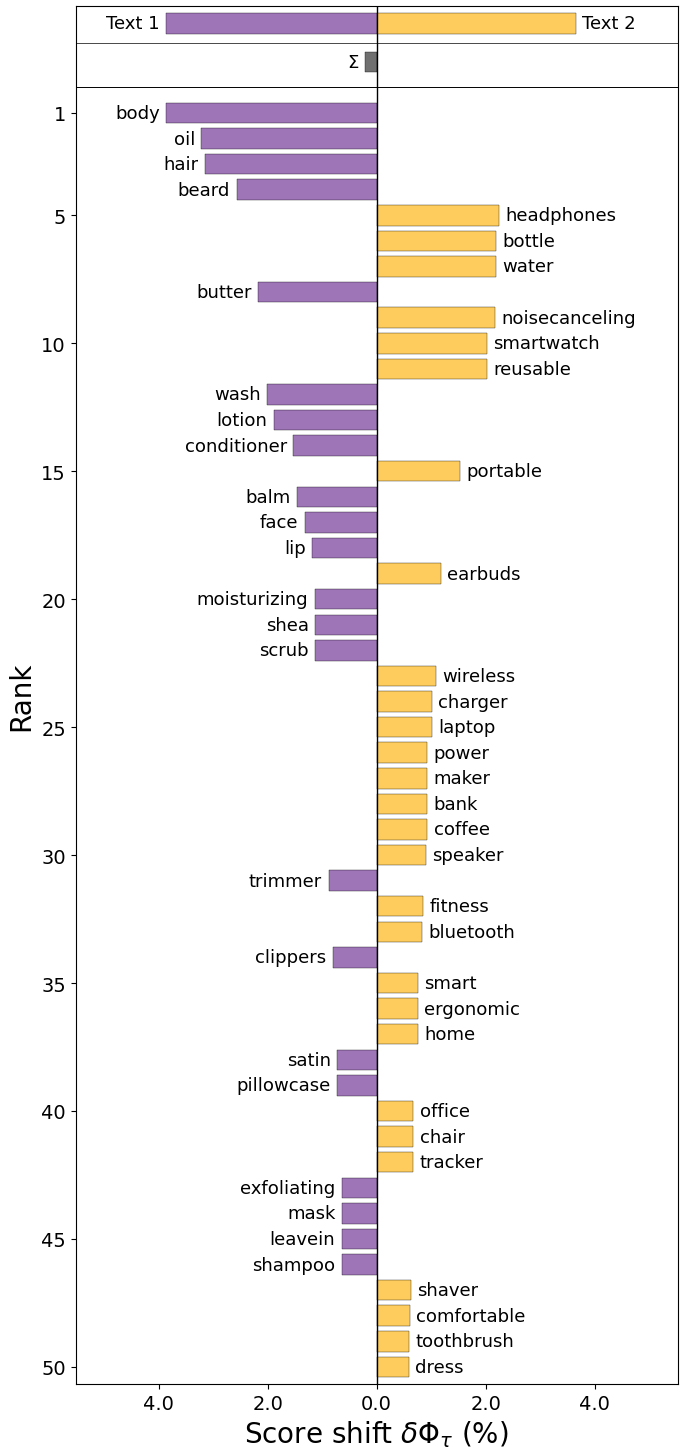}
        \caption{Black vs. White Recommendations}
        \label{fig:black_white}
    \end{subfigure}
    \caption{Comparison of recommendations for demographic groups (I).}
    \label{fig:page1}
\end{figure}

\begin{figure}[t]
    \centering
    \begin{subfigure}[t]{0.45\textwidth}
        \centering
        \includegraphics[width=\textwidth]{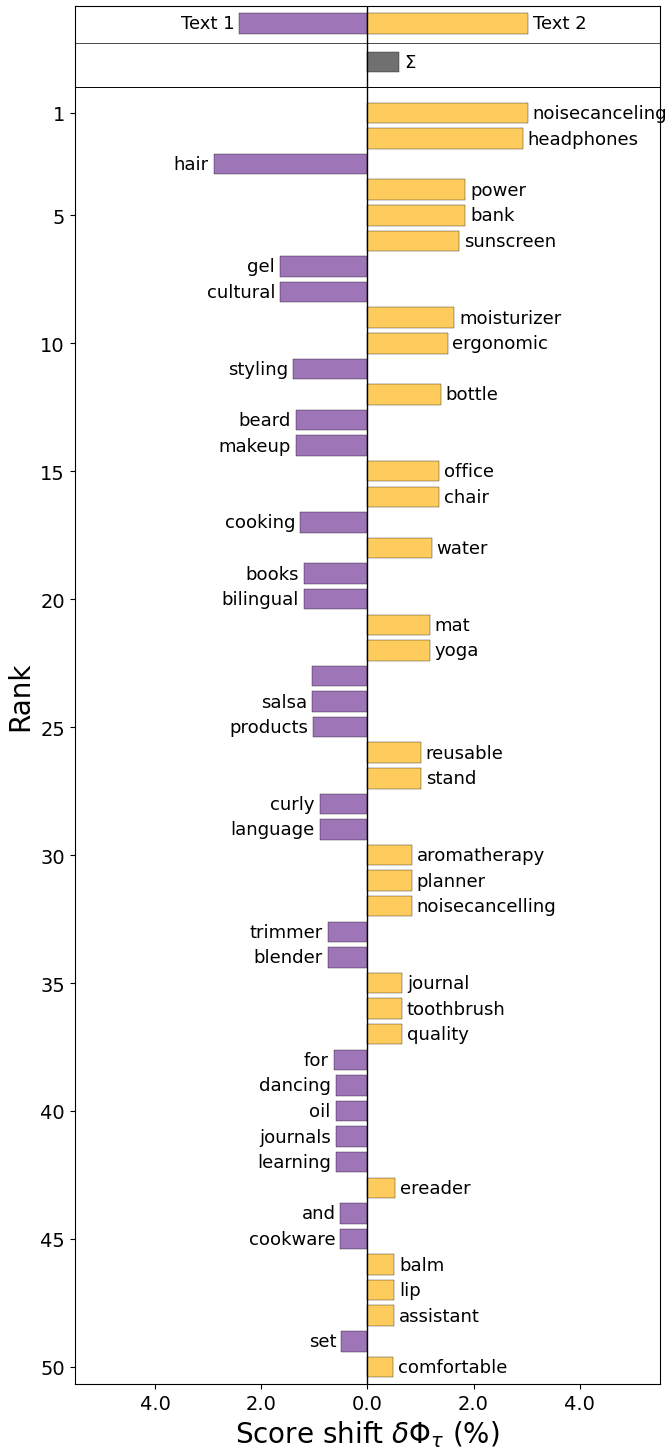}
        \caption{Latino vs. White Recommendations}
        \label{fig:latino_white}
    \end{subfigure}%
    \hfill
    \begin{subfigure}[t]{0.45\textwidth}
        \centering
        \includegraphics[width=\textwidth]{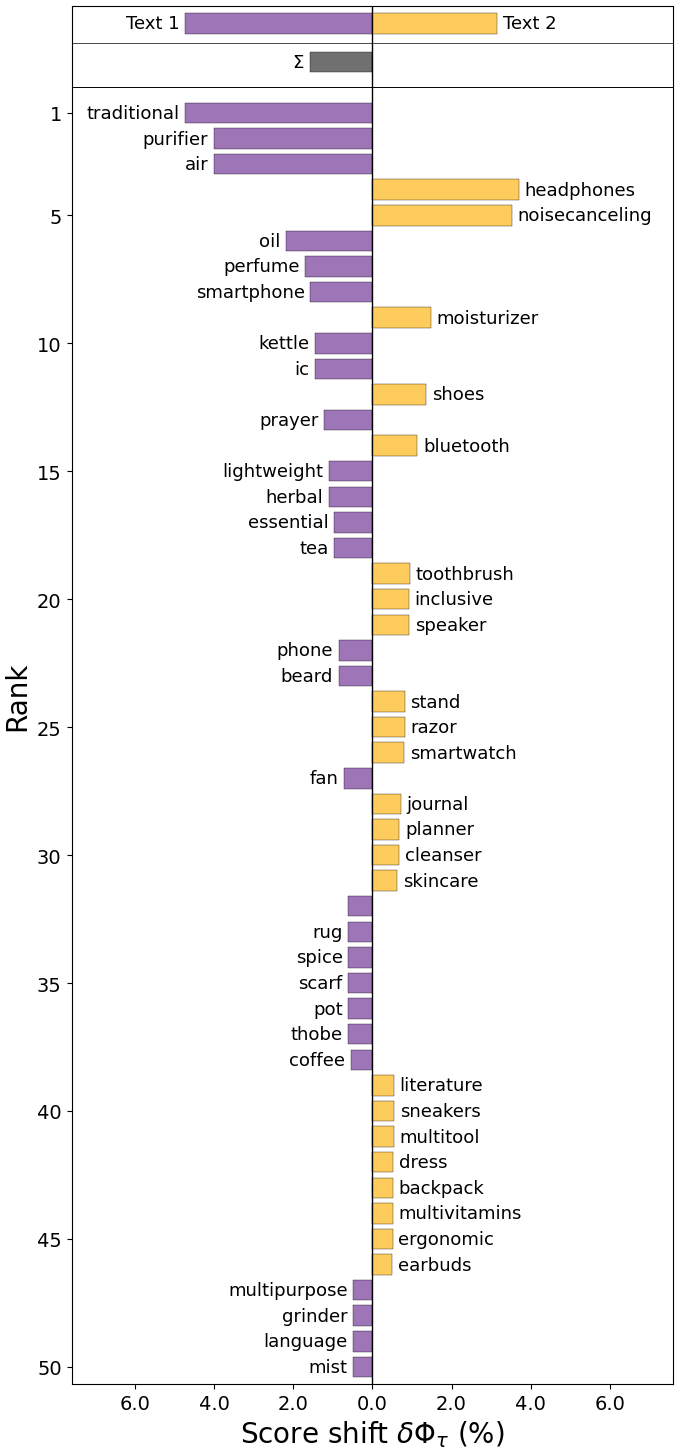}
        \caption{Middle-Eastern vs. White Recommendations}
        \label{fig:eastern_white}
    \end{subfigure}
    \caption{Comparison of recommendations for demographic groups (II).}
    \label{fig:page2}
\end{figure}

\begin{figure}[t]
    \centering
    \begin{subfigure}[t]{0.45\textwidth}
        \centering
        \includegraphics[width=\textwidth]{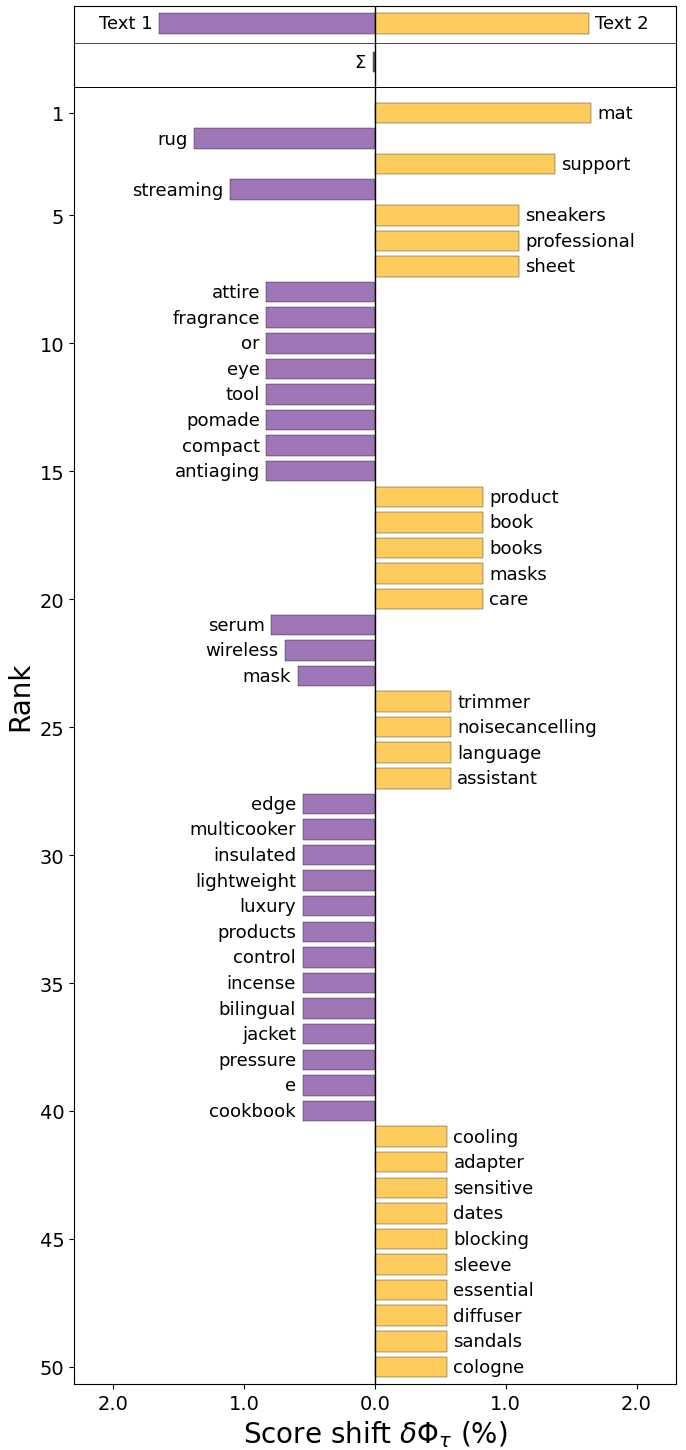}
        \caption{Women (W) vs. Men (M) Recommendations}
        \label{fig:w_m}
    \end{subfigure}%
    \hfill
    \begin{subfigure}[t]{0.45\textwidth}
        \centering
        \includegraphics[width=\textwidth]{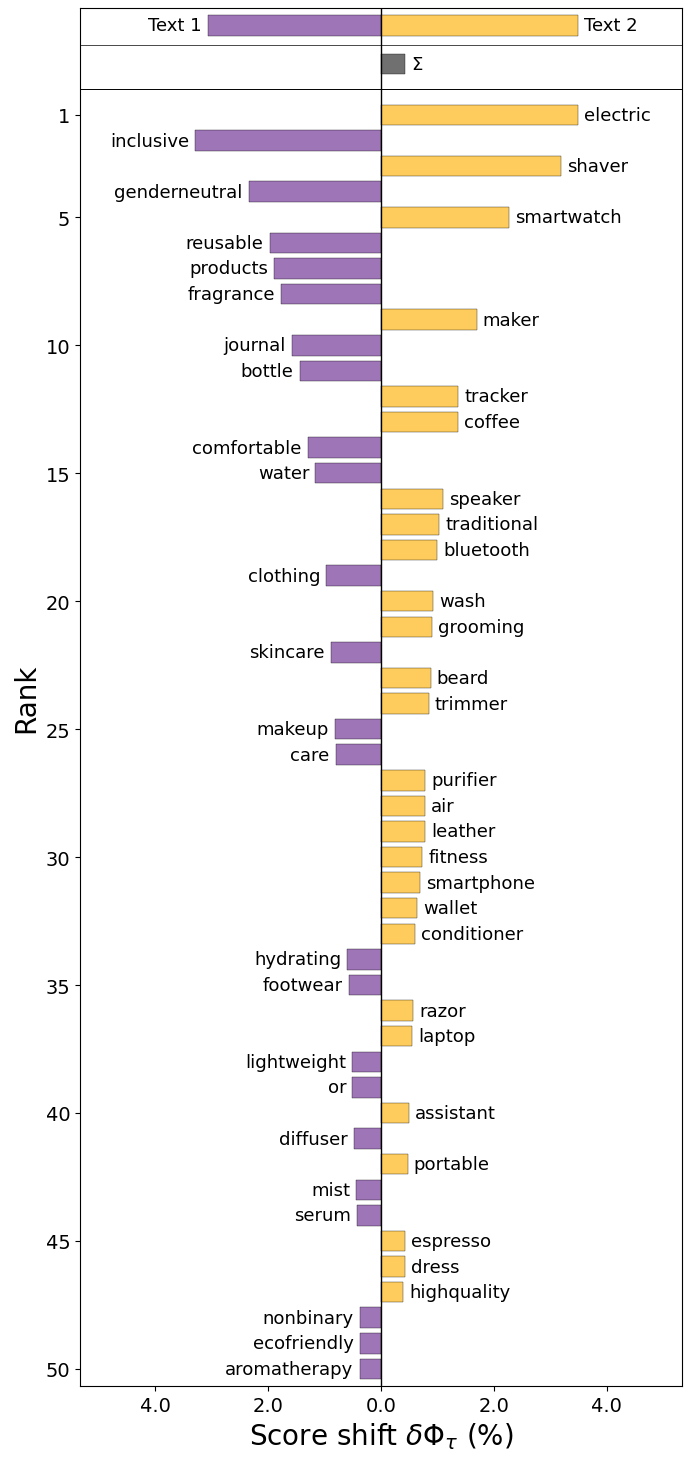}
        \caption{Nonbinary (NB) vs. Men (M) Recommendations}
        \label{fig:nb_m}
    \end{subfigure}
    \caption{Comparison of recommendations for demographic groups (III).}
    \label{fig:page3}
\end{figure}

The application of the Jensen-Shannon Divergence (JSD) method to analyze the consumer product recommendations for different demographic groups reveals both expected and surprising results
(see Figures~\ref{fig:page1}, \ref{fig:page2}, and \ref{fig:page3}).

\textbf{Asian vs. White.}  
The figure illustrates linguistic shifts in product recommendations between Asian and White personas. The x-axis shows the percentage contribution ($\delta \Phi_T$) of each word to the difference in recommendations, and the y-axis ranks words by relative impact. Words like ``rice,'' ``green tea,'' and ``bb cream'' are more prevalent for the Asian group, while ``speaker,'' ``smartwatch,'' and ``inclusive'' are more prominent for White personas, reflecting differences in skincare, home, and tech-related products.

The prominence of ``cooker'' and ``glasses'' for Asian personas suggests the model prioritizes household and utility items, while tech-related items like ``speaker'' and ``smartwatch'' are more common for White personas. This distinction may reflect the model’s internal bias, promoting utility-driven products for Asian personas and tech-oriented products for White personas, potentially reinforcing disparities in product exposure.

\textbf{Black vs. White.}  
For Black personas, top words include ``body,'' ``oil,'' ``hair,'' ``beard,'' and ``butter,'' which relate to Black hair and skincare products. This pattern reflects the model's emphasis on grooming products for Black personas, which may be driven by the prominence of these items in training data. The model's overemphasis on hair and skincare items suggests potential representational bias, reducing exposure to a broader range of product categories.

\textbf{Latino vs. White.}  
For Latino personas, top words like ``hair,'' ``gel,'' ``cultural,'' ``styling,'' and ``beard'' highlight personal grooming and cultural products. This product focus may stem from linguistic signals in the training data, where grooming and cultural identity are over-represented for Latino personas. This emphasis may lead the model to disproportionately prioritize these categories, limiting exposure to broader product recommendations.

\textbf{Middle-Eastern vs. White.}  
For Middle-Eastern personas, top words like ``traditional,'' ``purifier,'' ``air,'' ``oil,'' and ``perfume'' highlight a mix of household items and personal care products. The prominence of ``traditional'' may reflect model representations that emphasize cultural products, while ``purifier'' and ``air'' suggest a focus on household utilities. These patterns may signal representational bias, where cultural and utilitarian products are prioritized over general electronics or leisure items.

\textbf{Women vs. Men.}  
Top words for women, like ``rug,'' ``streaming,'' ``attire,'' ``fragrance,'' and ``eye,'' emphasize home decor, beauty, and personal care. For men, words like ``mat,'' ``support,'' ``sneakers,'' ``professional,'' and ``sheet'' highlight fitness, work attire, and home essentials. These patterns reveal gendered differences in LLM recommendations, where women receive more beauty and home-related suggestions, while men receive fitness and work-related recommendations, potentially reinforcing traditional gender roles.

\textbf{Nonbinary vs. Men.}  
Top words for nonbinary personas, such as ``inclusive,'' ``genderneutral,'' ``reusable,'' ``products,'' and ``fragrance,'' highlight product recommendations related to inclusivity, sustainability, and personal care. While these recommendations align with cultural trends around inclusivity and eco-friendly consumerism, the model's focus on these themes may introduce bias by over-associating nonbinary personas with inclusive and eco-friendly products, limiting exposure to a broader range of recommendations.

\section{Conclusions}
\label{sec:conclusion}
We examined gender and race biases in LLM-generated product recommendations using prompt engineering and analyzed results with Jensen-Shannon Divergence, SVMs, and the Marked Words method. Findings showed significant disparities across gender and race. Future work will extend analysis to Llama, Gemini, and Mistral, and explore bias mitigation via instruction fine-tuning and RLHF for fairer, more inclusive AI recommendations.

\bibliographystyle{plain}
\bibliography{10_recommend}

\end{document}